\documentclass[technote,compsoc]{IEEEtran}

\usepackage{graphicx}
\usepackage{subfig}
\usepackage{caption}
\usepackage{amssymb}
\usepackage{pifont}
\usepackage{amsfonts}
\usepackage{amsmath}
\usepackage{booktabs}
\usepackage{tabularx}
\usepackage{multirow}
\usepackage{xcolor,colortbl}
\usepackage{color}

\ifCLASSOPTIONcompsoc
  \usepackage[nocompress]{cite}
\else
  \usepackage{cite}
\fi

\definecolor{Gray}{gray}{0.85}
\definecolor{mycolor}{HTML}{00D1B2}
\newcommand{\rbf}[1]{\textbf{\textcolor{red}{#1}}}

\usepackage{array}
\newcolumntype{x}[1]{>{\centering\arraybackslash\hspace{0pt}}p{#1}}

\newcolumntype{a}{>{\columncolor{Gray}}c}
\newcommand{\rll}[1]{\textcolor{black}{#1}}

\def\etal{\textit{et al}.}
\def\ie{\textit{i.e.}}
\def\eg{\textit{e.g.}}

\usepackage[pagebackref=true,breaklinks=true,letterpaper=true,colorlinks,bookmarks=false]{hyperref}

\begin{document}

\title{Glance to Count: Learning to Rank with Anchors for Weakly-supervised Crowd Counting}

\author{\normalsize{Zheng Xiong, 
        Liangyu Chai,
	    Wenxi Liu,
	    Yongtuo Liu,
	    Sucheng Ren,
        and Shengfeng He,~\IEEEmembership{Senior Member,~IEEE}}
\IEEEcompsocitemizethanks{
\IEEEcompsocthanksitem Zheng Xiong, Liangyu Chai, Yongtuo Liu, Sucheng Ren, and Shengfeng He are with the School of Computer Science and Engineering, South China University of Technology, Guangzhou, China. E-mail: pandadreamer21@gmail.com, icepoint1018@gmail.com, csmanlyt@mail.scut.edu.cn, oliverrensu@gmail.com, hesfe@scut.edu.cn.
\IEEEcompsocthanksitem Wenxi Liu is with the College of Mathematics and Computer Science, Fuzhou University, China. E-mail: wenxi.liu@hotmail.com.}
}

\markboth{IEEE Transactions on Neural Networks and Learning Systems}%
{Shell \MakeLowercase{\textit{Xiong et al.}}: Glance to Count: Learning to Rank with Anchors for Weakly-supervised Crowd Counting}

\IEEEtitleabstractindextext{%
\begin{abstract}
Crowd image is arguably one of the most laborious data to annotate. In this paper, we devote to reduce the massive demand of densely labeled crowd data, and propose a novel weakly-supervised setting, in which we leverage the binary ranking of two images with high-contrast crowd counts as training guidance. To enable training under this new setting, we convert the crowd count regression problem to a ranking potential prediction problem. In particular, we tailor a Siamese Ranking Network that predicts the potential scores of two images indicating the ordering of the counts. Hence, the ultimate goal is to assign appropriate potentials for all the crowd images to ensure their orderings obey the ranking labels. On the other hand, potentials reveal the relative crowd sizes but cannot yield an exact crowd count. We resolve this problem by introducing ``anchors'' during the inference stage. Concretely, anchors are a few images with count labels used for referencing the corresponding counts from potential scores by a simple linear mapping function. We conduct extensive experiments to study various combinations of supervision, and we show that the proposed method outperforms existing weakly-supervised methods without additional labeling effort by a large margin.
\end{abstract}

}

\maketitle

\IEEEdisplaynontitleabstractindextext
\IEEEpeerreviewmaketitle

\vspace{-5mm}{\section{Introduction}\label{sec:introduction}}\vspace{-2mm}
As an important research topic in computer vision, crowd counting that aims to automatically count the number of individuals in images has been widely applied in many areas, \eg, video surveillance, traffic estimation, and congestion control.
Most recent approaches \cite{zhang2015cross}, \cite{zhang2016single}, \cite{cao2018scale}, \cite{li2018csrnet} rely mainly on fully-supervised annotation for individuals in crowd (\ie, placing a dot at the center of each individual) to estimate crowd density. Yet, such annotation process is extremely time-consuming and laborious. Especially for extremely dense scenarios, it is almost senseless to manually label over-heaped dots just for the purpose of representing crowd density in a scene. Such a tedious annotation process hinders the scale and diversity of crowd datasets and thus slows down the development of this area.

Recent work~\cite{yang2020weakly} revisits the regression-based counting method that ignores the exact individual locations and directly maps a crowd image to its crowd counts. However, the problem of annotation remain unsolved, as ground-truth crowd counts are required for training, which thus cannot prevent annotators from strenuously pinpointing each individual in the images.
Besides, some approaches~\cite{fiaschi2012learning}, \cite{arteta2014interactive} aim to bypass dense annotations with alternative interactions. Considering each annotated object in an image is atomic and equivalent, they require a few individual annotations instead of accurate locations of all objects. These methods are promising to relieve the annotation efforts while still achieving good performance, but they do not radically solve the problem.

\begin{figure}[t]
	\centering
	\setlength{\tabcolsep}{1pt}
	\begin{tabular}{>{\centering\arraybackslash\hspace{0pt}}p{.32\linewidth}
			>{\centering\arraybackslash\hspace{0pt}}p{.32\linewidth}
			>{\centering\arraybackslash\hspace{0pt}}p{.32\linewidth}}
		\includegraphics[width=\linewidth]{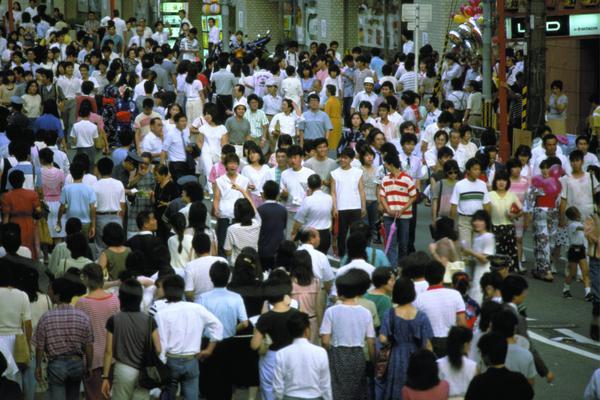}&
		\includegraphics[width=\linewidth]{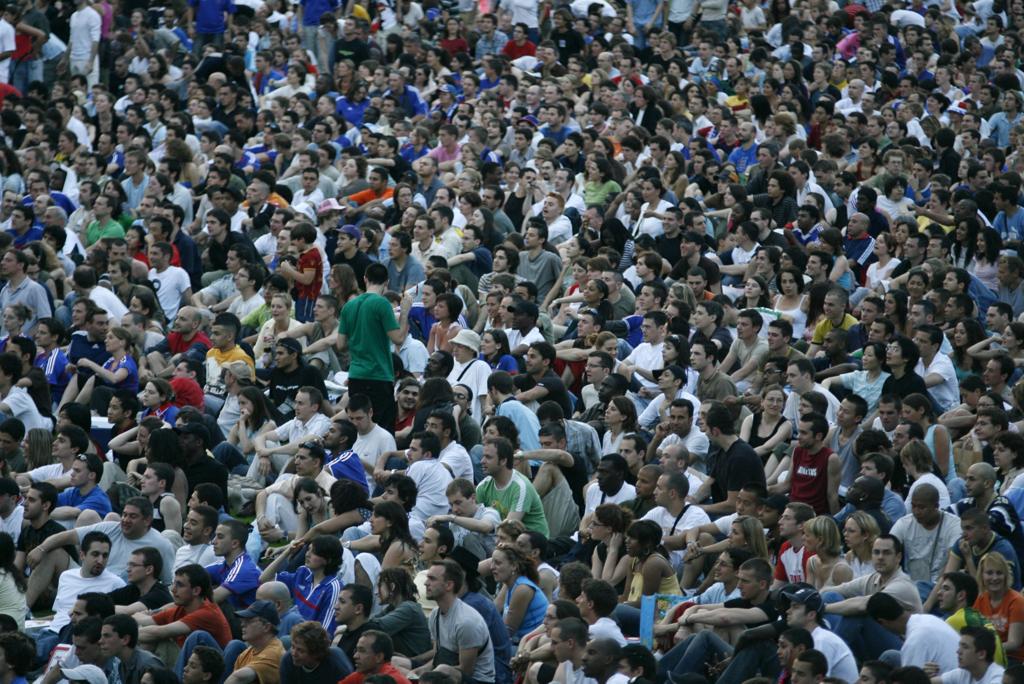}&
		\includegraphics[width=\linewidth]{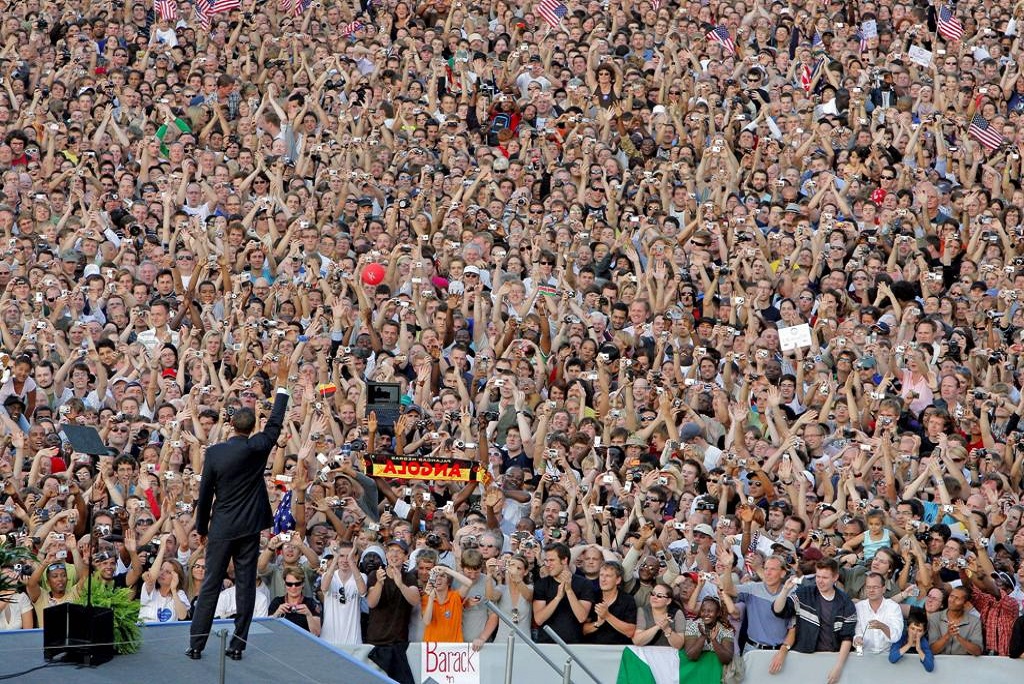}\\
		\small{$\mathcal{A}$} & \small{$\mathcal{B}$} & \small{$\mathcal{C}$}
	\end{tabular}\vspace{-2mm}
	\caption{For the crowd images with more than twice the number of differences, human can readily tell which one is of more people. We can assert that the number of people is: $count(\mathcal{A}) < count(\mathcal{B}) < count(\mathcal{C})$ intuitively. (The exact numbers: $count(\mathcal{A}) = 254$, $count(\mathcal{B}) = 580$, $ count(\mathcal{C}) = 1202$.  Please zoom in for a better view.) We aim to resolve the crowd counting problem by solely relying on ranking two images with high-contrast in crowd counts.}\vspace{-5mm}
	\label{fig:intuition}
\end{figure}

To address the above concern, we rethink the way of governing crowd counting models.
Intuitively, directly estimating the crowd count in an image is a challenging task even for a human expert.
But it is much easier to sense the relative density for a few crowd images with great contrast in population sizes. For example, for the three crowd images in Fig.~\ref{fig:intuition}, we can easily tell which one has the largest scale crowd.
Thus, this observation sheds light on a novel methodology of supervising the crowd counting models using \textit{ranking labels} for any pair of crowd images, where each ranking label indicates which one of the image pair contains more persons. Comparing to the existing annotation process, annotating ranking labels for images with large contrast in crowd sizes is an almost effortless task for human annotators, so establishing larger and more diverse crowd datasets becomes possible, which will be of great value to the community.
To this end, it boils down to the problem on how to effectively leverage the ranking labeled image pairs to supervise crowd counting models in such a weakly-supervised manner.

To exploit the ranking labels of crowd images, we propose a novel Siamese Ranking Network (SRN). Specifically, instead of directly estimating the crowd size of each image, a pair of crowd images are fed into two separate yet weight-sharing deep networks to predict their potentials that indicate the ordering of the counts in crowd images. For instance, in Fig.~\ref{fig:intuition}, the potential of $\mathcal{C}$ should be higher than those of $\mathcal{A}$ and $\mathcal{B}$. During the training phase, the magnitude relationship between a pair of predicted potentials can be supervised by the corresponding ranking labels of the crowd image pair. To this end, the purpose of model optimization is to assign appropriate potentials for all crowd images, which makes their orderings of potentials consistent with the ranking labels. Thus, the predicted potentials will be innately and positively correlated with the number of objects in crowd images, so they can be further applied to regress the absolute crowd counts. We construct a large amount of ranking labels corresponded to pairs of crowd images as the training set for training our proposed Siamese network.

However, the learned Siamese network can only produce potential scores as no crowd count labels are involved in training. To combat this problem, in the inference stage, we leverage a few images with crowd count labels as the reference, denoted as \textit{counting anchor set}, to disentangle the relationship between potential scores and actual crowd counts. Thanks to the positive correlation between the potentials and the ground-truth crowd counts, we establish a linear mapping function to fit the potential and counts of anchors, and we apply it to estimate the crowd count of the query image.
Our proposed annotation scheme is far more facile than the standard point-based annotation, especially in challenging crowd scenes, but we achieve comparable performances to those trained with a stronger supervision. Extensive experiments study various combinations of supervision in training, and we show that our model is superior to state-of-the-art weakly-supervised methods, even better than those trained with point-based labels.

Our contributions are three-fold:
\begin{itemize}
	\item We introduce a novel weakly-supervised crowd counting setting, which can reduce labeling cost largely and be notably beneficial to dense and congested crowds.
	\item We propose a simple but effective Siamese-training method and utilize the anchoring mechanism to estimate crowd counts and we verify its effectiveness in this task.
	\item Extensive experiments conducted on several challenging benchmarks study various combinations of supervision. We demonstrate that our method outperforms state-of-the-art weakly-supervised methods without any extra labeling effort.
\end{itemize} 
\vspace{-2mm}\section{Related Work}\label{sec:related-work}\vspace{-2mm}
\textbf{Fully-supervised Crowd Counting.}
In recent years, deep learning based method \cite{walach2016learning}, \cite{zhang2016single}, \cite{xiong2017spatiotemporal}, \cite{liu2018decidenet}, \cite{li2018csrnet} have attracted much attention in computer vision for crowd counting. The crowd counting methods mainly include detection-based methods and regression-based methods.
For the detection-based methods, Stewart \etal \cite{stewart2016end} propose to learn person detector relying on bounding box annotations to count. \cite{liu2019point} only requires point supervision to detect the human heads and count them in crowds simultaneously.
However, it is difficult to accurately detect heads or bodies in extremely dense and congested crowd scenes, and that always degrades counting performance.

Therefore, the mainstream idea is to train deep CNN networks for density regression. CNN-based regression methods learn a mapping from semantic features to density map and predict the total count. The main issue of regression-based counting task is the huge variation of instance scales. To tackle scale variations, employing multiple receptive fields is effective to learn from people with various sizes. For instance, several works \cite{zhang2015cross}, \cite{zhang2016single}, \cite{sam2017switching}, \cite{sindagi2017generating}, \cite{deb2018aggregated}, \cite{sam2018top} employ multi-column networks to obtain local or global contextual features to handle scale variations. \cite{cao2018scale}, \cite{liu2019adcrowdnet}, \cite{jiang2019crowd} utilize inception blocks to acquire different receptive fields. Several approaches \cite{li2018csrnet}, \cite{chen2019scale} combine the semantic features with dilated convolution for density estimation. Meanwhile, some works \cite{ranjan2018iterative}, \cite{zhao2019leveraging}, \cite{zhang2019relational}, \cite{wan2019adaptive} introduce attention mechanism which is effective in extracting foreground features. Considering performance gain from extra supervision, perspective maps \cite{liu2019context}, \cite{shi2019revisiting}, \cite{yan2019perspective} and depth maps \cite{lian2019density} are delivered to bring more scale guidance. On the other hand, combining with high-level tasks, \ie, localization \cite{ma2019bayesian}, \cite{liu2019recurrent}, segmentation \cite{shi2019counting}, depth prediction \cite{zhao2019leveraging}, can provide more accurate location labels for density regression and boost the counting with extra semantic information.

However, all the above CNN-based methods require a large number of labels during training, and annotating the crowd counting dataset is a labor-intensive and time-consuming task.

\textbf{Weakly-/Un-/Self-Supervised Crowd Counting.}
There are some weakly, self or unsupervised counting methods proposed with the considerations of relieving the labeling burden.
In the weakly-supervised setting, most methods are regression-based and adopt the image-level count label as the weak supervision signal for training. Idrees \etal \cite{idrees2013multi} leverage Fourier Analysis as feature extraction mechanisms to predict total counts. The work in \cite{von2016gaussian} applies Gaussian process as a weakly-supervised solution for crowd counting. For the CNN-based methods, \cite{yang2020weakly} proposes a soft-label sorting network to strengthen the supervision on crowd numbers beyond the original counting network, and \cite{lei2021towards} is a semi-supervised method combining a few location-level labels with count-level annotations. \cite{meng2021spatial} focuses on high confident regions while addressing the noisy supervision from unlabeled data as well in a semi-supervised manner. Although the above count-level methods are ``weakly-supervised'', the time spent for annotation actually is not remarkably reduced, while just the number of labels is reduced. Besides, Sam \etal \cite{sam2019almost} develop an autoencoder to achieve crowd counting under an almost unsupervised manner, and only few parameters are updated when training. By matching statistics of the distribution of labels, they propose a completely self-supervised training paradigm without using any annotated image in \cite{sam2020completely}. But the performance of unsupervised methods still exists a large gap with fully-supervised works. Besides, it is known that deep CNN-based crowd counting methods usually struggle with the overfitting problem due to existing small datasets and limited variety of them. To ease the overfitting problem, Wang \etal \cite{wang2019learning} explore generating synthetic crowd images to reduce the burden of annotation and alleviate overfitting.

Compared with the prior weakly-supervised methods, our proposed training settings based on ranking labels can effectively reduce the labeling burden, while maintaining the state-of-the-art performance.

\textbf{Learning to rank.}
Different from the standard machine learning tasks of regression or classification, the ranking tasks are not with precise ground-truth metric targets or class labels as supervision. These works are designed to handle with ordered ranks, \ie, which may be from human preferences, to predict the ordinal rank or relevant metric. There are many learning-to-rank works proposed in the literature. Some of the approaches, such as Ranking SVM \cite{herbrich2000large}, RankBoost \cite{freund2003efficient}, and RankNet \cite{burges2005learning}, focus on a pair of instances to learn their ranking function by minimizing the loss functions. \cite{sculleylarge} applies learning-to-rank to large-scale datasets by adapting the Stochastic Gradient Descent (SGD) method.

In contrast to the above works for predicting ranks, additional numerical references or other auxiliary supervision is required if we will make a prediction on the number of count.
Liu \etal \cite{liu2018leveraging} propose a learning-to-rank self-supervised strategy for utilizing available unlabeled images. And they show the ranking can be used as a proxy task for some regression tasks to solve the problem of limited size in existing datasets. However, relying the ranking of cropped images, which are cropped from the same source image, suffers the influence of similar paired patterns. 

In this paper, the proposed ranking strategy works on pairwise instances of different images, similar to ranking SVM \cite{herbrich2000large}. The setting can be free from the aforementioned problems from the same source pattern. And our method alleviate both overfitting problem and intensive annotation burden. The performance in a weakly-supervised setting is comparable to location-level supervised methods, such that can be feasibly applied in practical applications.

\section{Proposed Method}\label{sec:method}
\subsection{Problem Formulation}

Given a set of crowd images $X = \{x_1, x_2, ... ,x_n\}$, where $x_i$ is the sample of crowd images, and $x_i \in \mathbb{R}^{H \times W \times C}$, where $H$, $W$ and $C$ denote the height, width, and number of channels respectively, our goal is to predict the object count $y_i$ of a crowd image $x_i$.
The existing works achieve the goal by learning a mapping function $M: X \rightarrow Y$ to predict the labels, \ie, $Y=\{y_1, y_2, ... , y_n\}$. Different from the previous practice, we utilize \textit{ranking labels} for supervision instead of count labels. Count labels refer to the object number of crowd images, while ranking labels are binary (i.e. $\{-1, 1\}$), each of which indicates the size relationship between the counts of two crowd images.

Formally, the crowd dataset $X$ can be expanded as a set $P$ of ranking labeled pairs. Each element of the ranking pair set $P$ in the crowd dataset $X$ is denoted as a tuple $\langle x_i, x_j, q_{i,j} \rangle, 1 \le i ~\textless~ j \le n$, where the binary ranking label $q \in \{-1,1\}$. When the crowd counts of $x_i$ and $x_j$ are subject to $y_i \gg y_j$, their size relationship can be easily identified, which means that the crowd image $x_i$ ranks higher than $x_j$, so that the corresponding label $q_{i,j}$ is set to $1$ and vice versa.

\rll{In favor of learning the ranking relationship among a set of crowd images, we introduce the \textit{crowd counting potential} $v$ that is positively correlated with the actual crowd count $y$. Thus, the goal is to learn an estimating function $f(x;w)$ to predict the potentials $v$ for all the crowd images with ranking labels, which aims to ensure the ordering of the assigned potentials obey the ranking labels. After that, the estimation function can be further used to predict the potential for any query image and the anchoring mechanism can map its potential to the absolute crowd count.}

\begin{figure*}[t]
	\centering
	\vskip -0.25cm
	\includegraphics[width=\textwidth]{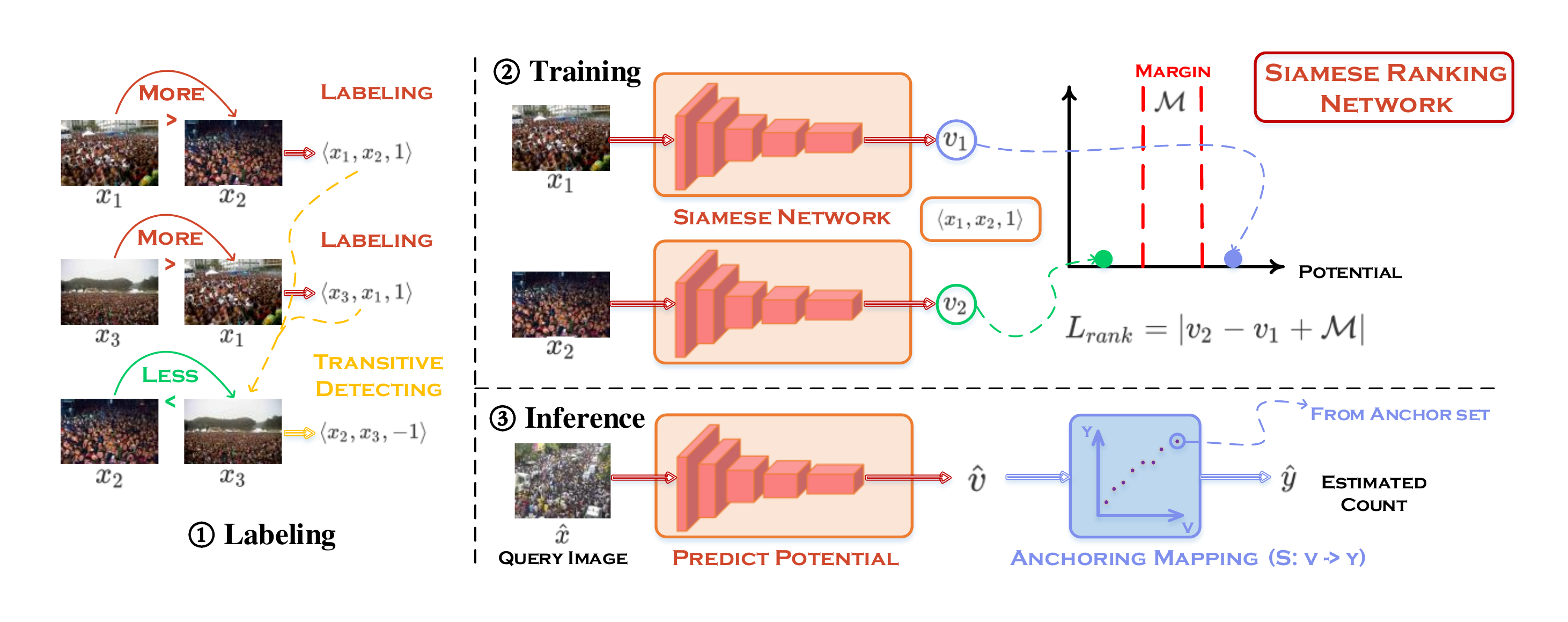}
	\vskip -7px
	\vspace{-2mm}\caption{{Our framework consists of three stages. (1) Constructing a DAG to detect transitive relation amongst crowd images, i.e., automatic labeling, which establishes a crowd dataset with ranking labels. (2) Training the Siamese Ranking Network on ranking labels to minimize the ranking hinge loss and predict potentials (e.g. $v_1,v_2$) for crowd images, where the parameter $\mathcal{M}$ denotes the margin of the loss. (3) On the inference stage, the count of a crowd image $\hat{x}$ can be estimated by mapping the predicted potential $\hat{v}$ to actual crowd count $\hat{y}$.}}\vspace{-2mm}
	\label{fig:architecture}
\end{figure*}

\subsection{Labeling Ranking Pairs}

For a pair of unlabeled crowd images $\{x_1, x_2\}$, if it is observed that the crowd number $y_1$ is definitely much greater than $y_2$, the ranking label $q_{i,j}$ is set to $1$, and vice versa.
If the relationship is hard to identify, no label will be added to the ranking pair set $P$.
Regarding various crowd scenarios, there are different strategies to label ranking pairs.

First, we can exploit the images from public benchmarks to train our model.
For most existing crowd counting benchmarks (\eg, ShanghaiTech \cite{zhang2016single}, UCF-QNRF \cite{idrees2018composition}), they can provide the dense annotations of object locations for crowd images and the count labels can thus be acquired without effort, which means we can easily convert them into ranking labels.
Generally, for crowd dataset $X$, the size of all ranking pairs $|P_{all}|$ is $O(n^2)$ (where $n$ is the size of dataset), in which any two images could be formed as a ranking pair. However, the number of ranking labels is much less than $n^2$, since we are only allowed to manually label the obviously distinguishable pairs of crowd images. In addition, considering that the ranking relationship is transitive, if the ranking labels, $\langle x_A, x_B, 1\rangle$ (i.e. $y_A > y_B$) and $\langle x_B, x_C, 1 \rangle$ (i.e. $y_B > y_C$), already exist in the annotated set, then $\langle x_A, x_C, 1\rangle$ or $\langle x_C, x_A, -1\rangle$ will be automatically added to the annotated set as shown in Fig.~\ref{fig:architecture}. Particularly, automatic labeling can be realized by detecting the connectivity in a directed acyclic graph, $G = (V, E)$. In specific, $V$ denotes the set of vertices, which includes all the images that occurred in the ranking pairs. $E$ denotes the set of arcs, where $\langle x_i, x_j, q_{i,j}\rangle, q_{i,j}=1$ refers to an arc $i \rightarrow j$ from the vertex $i$ to th vertex $j$ on the graph $G$ .

Nevertheless, the crowd images from public benchmarks are not sufficient to cover a large variety of scenes, so we also annotate the crowd images in the wild, so as to increase the diversity of the training data and thus strengthen the model generalizability. Thus, we attempt to collect as many unlabeled crowd images as possible. On the other hand, we restrict each collected unlabeled image to be ranked for a few times only. Actually, in practice, most of the collected images will be ranked for once, so that our crowd counting model can focus more on unseen images.

\subsection{Pairwise Ranking Model}
Our proposed Siamese ranking network architecture is illustrated in Fig.~\ref{fig:architecture}.
In specific, we employ a deep Siamese Network \cite{chopra2005learning} as the ranking model for pairwise crowd images, which is consisted of two branches of networks that share weights. The input of our model is a ranking pair $\{x_1, x_2\}$ from the set $P$. Each branch of the networks is fed with one of the image pair and outputs the corresponding potentials $v_1$ and $v_2$. We leverage a convolutional neural network (CNN) based crowd counting architecture, CSRNet \cite{li2018csrnet}, as the backbone of the Siamese network. The backbone is composed of a pre-trained CNN as the frontend for 2D feature extraction and a dilated convolutional layer for enlarging the reception fields on the backend.

To supervise the ranking labels, we propose to associate the predicted potentials of two network branches with the corresponding ranking labels.
Inspired by Ranking Support Vector Machine \cite{herbrich2000large}, we can train the network by minimizing the ranking hinge loss as follows:
\begin{align}
	\mathcal{L} &= \sum_{x_i,x_j}max(0, f(x_j; w) - f(x_i; w) - \mathcal{M}),\nonumber\\
	&\text{s.t. }	 \langle x_i, x_j, 1\rangle \text{ or } \langle x_j, x_i, -1\rangle \in P,
\end{align}
where $\mathcal{M}$ is a hyper-parameter that indicates \rll{a margin to maintain the potential difference between paired images} and $w$ refers to the network weights.

\rll{Specifically, during training, we want the model to discard the redundant samples that are easy to discriminate and cannot contribute to the model optimization.
	To do so, we set a \textit{hard sample filter} which requires the potentials $v_i$ and $v_j$, inferred from $\{x_i,x_j\}$, should be subject to the condition $\frac{v_i}{v_j} < \xi$ (or $\frac{v_j}{v_i} < \xi$), where $\xi$ is a predefined threshold to determine if the ranking labels should be abandoned.}

\subsection{Anchoring Mechanism}
\rll{During inference, for any query crowd image $\hat{x}$, its predicted potential can be used to regress its crowd count. Thus, it is intuitive to compare the potential $\hat{v}$ of the query image (i.e., $\hat{v} = f(\hat{x})$) against a set of exemplar crowd images with their ground-truth counts known. }

\rll{In particular, we introduce a counting anchoring mechanism. We denote the labeled exemplar images as \textit{counting anchor set}. The anchor set includes a set of crowd images sampled from the training set, and their counts are specifically annotated and distributed over a large counting range, \eg, 30 $\sim$ 3,000. Thus, we can estimate the scaling mapping function $S:v \rightarrow y$ that projects the predicted potentials of the images in the anchor set to crowd counts via linear regression, as shown in Fig.~\ref{fig:architecture} \ding{174}.}


\subsection{\rll{Ranking v.s. Regression}}
The reasons of choosing ranking-based scheme over regression-based methods are two fold. First, comparing to the point-based annotation, ranking labels are much easier to obtain, i.e., annotators only need to label ``close'', ``less than'', and ``greater than'' for any two crowd images. Second, ranking problem is simpler to solve than regression-based methods, because ranking relationship are invariant to any image-level geometrical transformation, which can thus improve the model robustness. On the contrary, optimizing regression model is difficult and tends to get stuck with local-minima.
Although it is easy to obtain the optimal solution for the ranking based formulation, it may not always be optimal for crowd counting. This is because the theoretical upperbound of the ranking optimization objective is easy to approach. It seems that simply learning a good ranking model is insufficient for counting task, and a suboptimal regression model can also give perfect ranking. But a model with perfect ranking performance may yield poor regression performance.

Therefore, it is a trade-off for choosing ranking or regression. To further improve our model we propose to integrate regression and ranking-based schemes. To this end, crowd counting can be achieved in such a hybrid weakly-supervised setting. The optimization objective is consisted of two terms, i.e., pairwise ranking loss $L_{rank}(P; w)$ and regression loss $L_{reg}(D; w)$. In each iteration during training, there will be a crowd image randomly selected from $D$ and a ranking pair randomly selected from $P$ as network input. Formally, the hybrid optimization loss is defined as below:
\begin{align}
	\mathcal{L}^{++} = \min_{w} \mathcal{L}_{rank}(P; w) + \alpha \mathcal{L}_{reg}(D; w),
\end{align}
where $D$ is the dataset including crowd images with count label for regression, and the parameter $\alpha$ trades off between pairwise ranking loss and regression loss. Note that, with more crowd labels used in regression, the performance benefited from regression is greater.
To make full use of the supervision from counting anchor set, we can utilize these few anchor samples with count labels as $D$ for regression.
In the most extreme case, we can attach all the training images with count labels for optimizing the regression loss term, so that it transforms our formulation into a count-label based optimization completely. Yet, with the involvement of the ranking loss term, the performance is significantly superior to the pure regression-based methods.
\section{Experiments}\label{sec:application}
We conduct extensive experiments to evaluate our approach on several crowd benchmarks: ShanghaiTech PartA \cite{zhang2016single}, UCF-QNRF \cite{idrees2018composition}, UCF\_CC\_50 \cite{idrees2013multi}, and JHU-CROWD++ \cite{sindagi2019pushing}, \cite{sindagi2020jhu}.
We compare our approach against other weakly-supervised counting methods. Note that, compared to the existing weakly-supervised methods, our ranking-based model requires weaker supervision.
In this section, we first describe the implementation details and evaluation metrics. Then, we compare and evaluate our method with the peer weakly-supervised state-of-the-art methods. Last, we perform comprehensive ablation studies to delve into our model.

\subsection{Implementation Details and Metrics}
\textbf{Implementation Details.} We implement our model using Pytorch \cite{paszke2019pytorch}. The network backbone used in our experiment is CSRNet \cite{li2018csrnet}. Apparently, the backbone can also be replaced by other CNN-based counting model. The proposed network is trained using Adam solver \cite{kingma2014adam} as the optimizer with a mini-batch size of 1. The learning rate is set to 5e-5. Except for the ablation study, the margin of SVM in our methods is set to 0.5. Since the datasets include some high-resolution (HR) images, we downsample them to the resolution of less than $700 \times 800$.

\textbf{Evaluation Metrics.}
There are two metrics widely used to evaluate the performance of crowd counting. Mean Absolute Error (MAE) implies count estimation accuracy, which is formally defined as,
\begin{align}
MAE = \frac{1}{\mathcal{N}} \sum_{i=1}^{\mathcal{N}}|c_i - c_i^{GT}|,
\end{align}
where $\mathcal{N}$ is the number of images in the testing set. $c_i$ is the predicted count for $i$-th image, while its actual count is $c_i^{GT}$.
Mean Squared Error (MSE) is the metric for variance of counting estimation to reflect the robustness of prediction, which is defined as,
\begin{align}
MSE = \sqrt{\frac{1}{\mathcal{N}} \sum_{i=1}^{\mathcal{N}}(c_i - c_i^{GT})^2}.
\end{align}
\textbf{\rll{Anchoring Mapping.}}
\rll{During inference, to evaluate the proposed ranking method, the scaling mapping function $S:v\rightarrow y$ from potential scores to real count numbers can be learned by linearly fitting the images in the counting anchor set, where $\{v, y\}$ is paired and known.}

\begin{table*}[t]
	\centering
	\caption{Comparison of our proposed method with baselines and related methods on ShanghaiTech Part A \cite{zhang2016single} and UCF-QNRF \cite{idrees2018composition}. ``label level'' refers to the supervision level of training. \ding{52} means the model employs all the labels under the corresponding level of supervision, and \ding{70} means the model employs a few labels at this supervision level. $*$ indicates the $0.1\%$ of the parameters are tuned with location-level supervision. \rll{Note that, \textit{Ours++ (Fully)} exploits full count labels as supervision, and ranking labels can be auto-generated from count label without extra annotating effort.} The best results in weakly-supervised setting are highlighted in \rbf{red}.}
	\begin{tabular}{c||x{1.3cm}x{0.8cm}x{1.3cm}x{1.3cm}|x{0.95cm}x{0.9cm}|x{0.95cm}x{0.9cm}}
		\toprule
		&\multicolumn{4}{c|}{\textbf{Label level}} & \multicolumn{2}{c|}{\textbf{ST PartA}\cite{zhang2016single}} & \multicolumn{2}{c}{\textbf{UCF-QNRF}\cite{idrees2018composition}} \\
		\midrule
		\textbf{Method} & \textbf{Location} & \textbf{Count} & \textbf{Ranking} & \textbf{No label}& \textbf{MAE$\downarrow$} & \textbf{MSE$\downarrow$} & \textbf{MAE$\downarrow$} &\textbf{MSE$\downarrow$} \\
		\midrule
		MCNN \cite{zhang2016single} (2016) & \ding{52} &  &  &  & 110.2 & 173.2 & 277.0 & 426.0 \\
		Switching-CNN \cite{sam2017switching} (2017) & \ding{52} &  &  &  & 90.4 & 135.0 & 228.0 & 445.0\\
		CSRNet \cite{li2018csrnet} (2018) & \ding{52} &  &  &  & 68.2 & 115.0 & 119.2 & 211.4 \\
		CAN \cite{liu2019context} (2019) & \ding{52} &  &  &  & 62.3 & 100.0 & 107.0 & 183.0 \\
		ADSCNet \cite{bai2020adaptive} (2020) & \ding{52} &  &  &  & {55.4} & {97.7} & {71.3} & {132.5} \\
		\midrule
		GWTA-CCNN \cite{sam2019almost} (2019) & $*$ &  &  & \ding{52} & 154.7 & 229.4 & - & -\\
		CSS-CCNN \cite{sam2020completely} (2020) &  &  &  & \ding{52} & 207.3 & 310.1 & 442.4 & 721.6 \\
		IRAST (Label only) \cite{liu2020semi} (2020) & \ding{70} &  &  &  & 98.3 & 159.2 & 147.7 & 253.1  \\
		IRAST \cite{liu2020semi} (2020) & \ding{70} &  &  & \ding{52} & 86.9 & 148.9 & 135.6 & 233.4 \\
		CCLS \cite{yang2020weakly} (2020) &  & \ding{52} &  &  & 104.6 & 145.2 & - & - \\
		MATT \cite{lei2021towards} (2021) & \ding{70} & \ding{52} &  &  & 80.1 & 129.4 & - & - \\
		\midrule
		CSRNet(Count label only) &  & \ding{52} &  &  & 85.6 & 128.1 & 149.0 & 245.3 \\
		CSRNet(Count label only)+L2R \cite{liu2018leveraging} &  & \ding{52} &  &  & 84.91 & 125.2 & 144.6 & 238.2 \\
		Ours &  &  & \ding{52} &  & 93.4 & 142.5 & 165.3 & 277.7  \\
		Ours++ &  & \ding{70} & \ding{52} &  & 91.6 & 138.5 & 158.7 & 266.7  \\
		Ours++ (Fully) &  & \ding{52} & \ding{52} &  & \rbf{76.9} & \rbf{113.1} & \rbf{133.0} & \rbf{218.8} \\
		\bottomrule
	\end{tabular}
	\label{table:1}\vspace{-2mm}
\end{table*}

\subsection{Comparison with State-of-the-arts}
Compared with previous weakly-supervised counting methods, our supervision scheme is unique, which requires less supervision information. Here, we mainly compare the proposed methods with other approaches with diverse supervised settings. Our comparison evaluation is conducted on ShanghaiTech PartA \cite{zhang2016single} and UCF-QNRF \cite{idrees2018composition} datasets.

\textbf{ShanghaiTech Part A Dataset.}
ShanghaiTech Part A dataset \cite{zhang2016single} is a large-scale crowd counting dataset, which is composed of 482 images with 244,167 annotated persons. The training set includes 300 images with 162,707 annotated persons, and the remaining 182 images are for testing. The images are captured from the Internet and the number of the humans ranges from 33 to 3139 per image. Following the assumption of glance annotation, the available training ranking pairs are 24,386. Due to our transitive automatic labeling, the times of manually annotating ranking pairs is reduced to 16,194. And we randomly pick up 50 images to set up the counting anchor set, which corresponds to different crowd density levels in the training set. In total, the number of ranking pairs is only $1/10$ of location-based labels. Adding the labeling efforts of counting anchor set, the total labeling amount is around 1/4 of the original labels on ShanghaiTech Part A dataset.

\textbf{UCF-QNRF Dataset.}
The UCF-QNRF dataset \cite{idrees2018composition} contains 1,535 images with counts varying from 49 to 12865 including 1,251,642 annotated heads, thus the average count is around 815 per image. The training set includes 1,201 images, and 334 images are for testing. We also randomly pick up 70 images from different range of density as the counting anchor set. Owing to the huge data size, the number of available ranking pairs is exploded, thus it is intractable to annotate all pairs in the real world. To handle the problem, the hyper-parameter $\mathcal{N}_{sim}$ is provided to simulate the number of annotated ranking pairs when training for ranking. Here we set $\mathcal{N}_{sim} = 48,000$, which means the given set $P$ with size 48,000 is fixed before training and these pairs are available training samples.

\textbf{Categories of Crowd Counting Methods.}
Generally, the supervision of the evaluation methods can be roughly categorized from laborious to effortless as four levels, location level, count level, ranking level and no label:
\begin{itemize}
	\item \textit{Location level} supervision relies on location-based density maps as the optimizing objective. For a dense-scene crowd image, it requires a lot of effort to complete the annotation of hundreds of locations.
	\item \textit{Count level} supervision is based on crowd count numbers without the location supervision. There are not many labels but one-by-one manual counting is required.
	\item \textit{Ranking level} supervision is a subjective ranking label for pair-wise images and the label can be conducted only at a glance.
	\item \textit{No label} supervision is not to use any annotated label, only use raw crowd images for input.
\end{itemize}

\textbf{Experimental Results.} In experiments, we apply several variants of our proposed method for comparison. Particularly, the model trained only with ranking labels is denoted as \textit{Ours}, whereas \textit{Ours++} refers to the model that is trained with the ranking and cost-free regression labels from the counting anchor set. Besides, \textit{Ours++ (Fully)} attaches all training images with count label for regression along with ranking label for ranking.

The quantitative comparison with the state-of-the-art methods on these two datasets is presented in Table~\ref{table:1}. The fully-supervised location-level methods are listed in the first part of table, and the weakly-/semi-/un-supervised methods for comparision are listed in the second part. The performance of the proposed method and compared baselines are at the bottom. \textit{CSRNet (Count label only)} represents the regression-based baseline supervised only by count labels. \textit{CSRNet (Count label only) + L2R \cite{liu2018leveraging}} is to add "cropping rank" supervision for the above label-only model. Note that these comparative weakly-supervised methods, \eg, IRAST \cite{liu2020semi}, CCLS \cite{yang2020weakly}, and our proposed method are also based on the same feature extractor, CSRNet \cite{li2018csrnet}, for the fair comparison. As observed, the proposed method outperforms the other competing methods with similar supervision level on the benchmarks.

\begin{figure}[t]
	\centering
	\setlength{\tabcolsep}{1pt}
	\begin{tabular}{>{\centering\arraybackslash\hspace{0pt}}p{.040\linewidth}
			>{\centering\arraybackslash\hspace{0pt}}p{.3\linewidth}
			>{\centering\arraybackslash\hspace{0pt}}p{.3\linewidth}
			>{\centering\arraybackslash\hspace{0pt}}p{.3\linewidth}}

		\rotatebox{90}{\small{\centering{ST PartA}}}&
		\includegraphics[width=\linewidth]{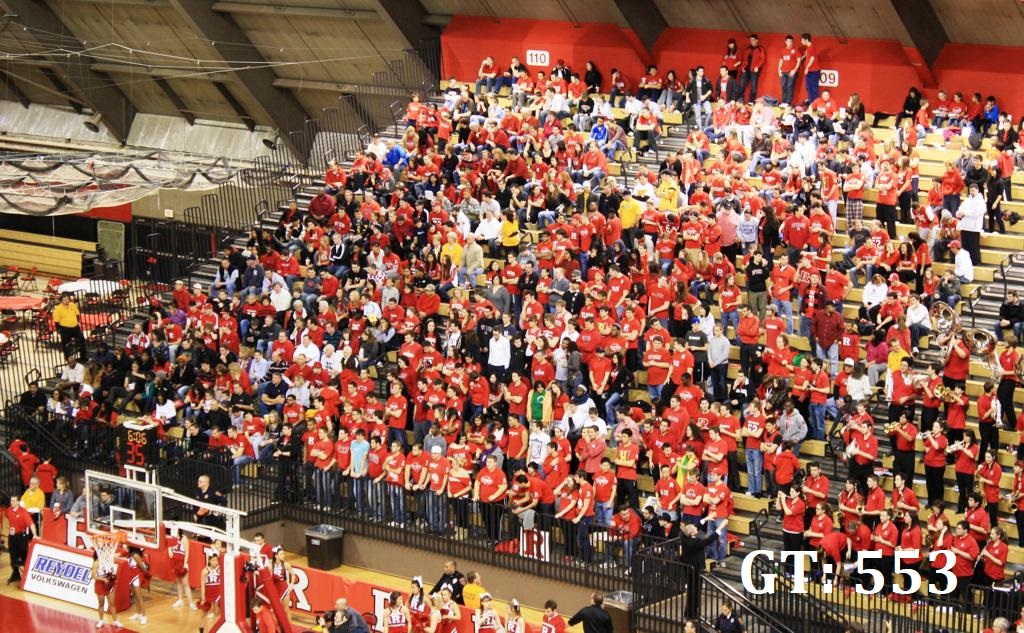}&
		\includegraphics[width=\linewidth]{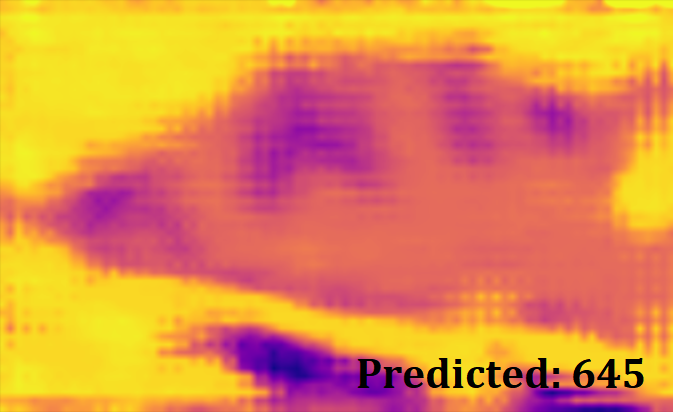}&
		\includegraphics[width=\linewidth]{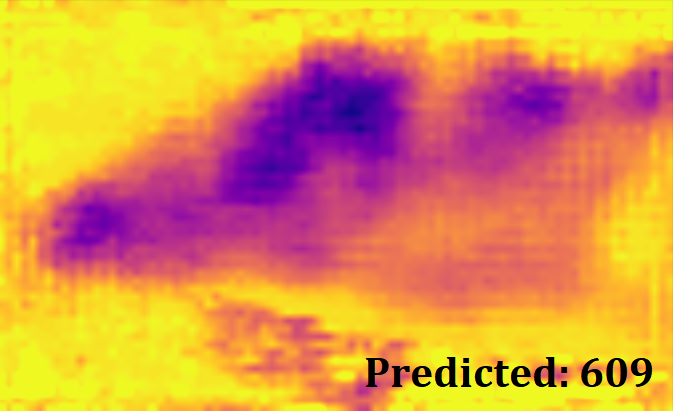}\\
		
		\rotatebox{90}{\small{\centering{UCF-QNRF}}}&
		\includegraphics[width=\linewidth]{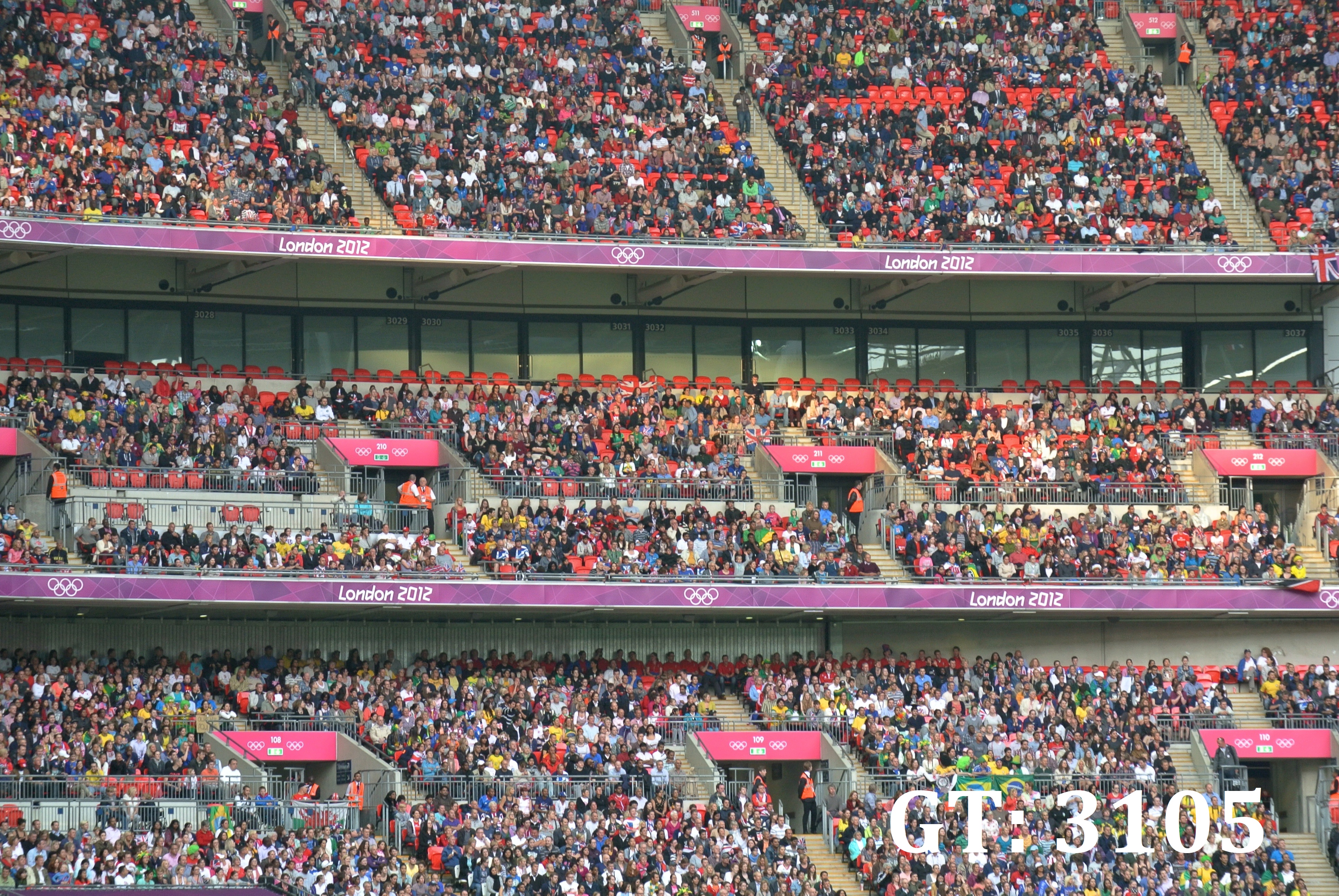}&
		\includegraphics[width=\linewidth]{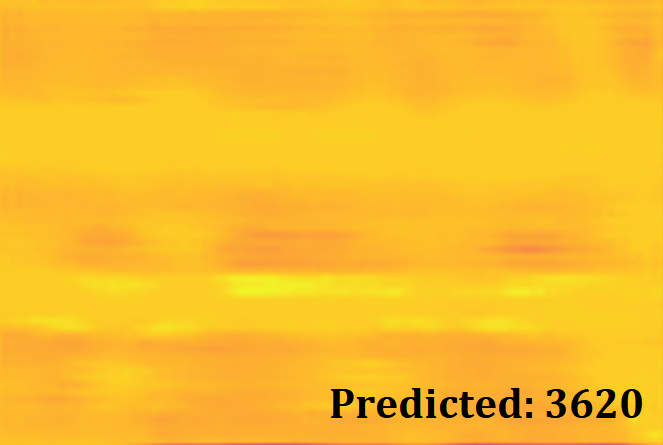}&
		\includegraphics[width=\linewidth]{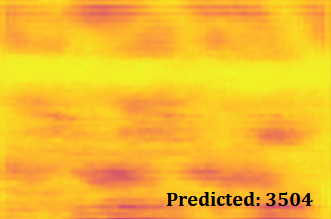}\\
		&Input & Ours & Ours++ (Fully)
	\end{tabular}
    \caption{Qualitative visualization of predicted density maps on two examples from ShanghaiTech PartA and UCF-QNRF by our methods. It is shown that ranking-based methods are quite good at distinguishing the crowd regions.}\vspace{-2mm}
	\label{fig:visualize}
\end{figure}

As shown in Table~\ref{table:1}, it is clear that our method outperforms the weakly-supervised regression method, CCLS \cite{yang2020weakly}, the state-of-the-art method with the same supervision level and our annotating workload is less for \textit{Ours, Ours++}. \textit{Ours++ (Fully)} achieves much less counting error than the label-only baseline due to the ranking pair supervision, and also achieves less error than the L2R supervised count label regression model. Thus, it demonstrates that ranking labels are of great use for regression at the count level supervision. Moreover, the performance of \textit{Ours++ (Fully)} is the best in weakly-supervised setting on MAE and MSE, and even close to location level baseline, CSRNet \cite{li2018csrnet}.
We visualize the density maps delivered from the last layer from network as illustrated in Fig~\ref{fig:visualize}. Although there is no location points as supervisory signals, the estimation of our approach are close to the ground truth density maps, which shows the capability of our model to identify individuals in crowds.

\subsection{Ablation Study}

\textbf{Impact of the margin on SVM.} We set hyper-parameter margin $\mathcal{M}$ to 0.5 in the previous experiments. To investigate the impact of margin, we conduct experiments on ShanghaiTech part A datasets. As shown in Table~\ref{table:2} (left), the results demonstrate setting different margin of SVM does not affect the counting performance significantly. The crowd counting performance are similar to our current results when margin is set properly, which means the proposed method is robust.

\textbf{The size of the counting anchor set.} So far, we conduct our experiment with the size of the counting anchor set $50$ (\textit{Ours++} and $300$ for \textit{Ours++(Fully)}, which is the full of ShanghaiTech Part A training set. It is unclear the performance gain is affected with varying its size during training phase. To investigate this, we conduct an experiment that applies different sizes on ShanghaiTech Part A. The results are shown in Table~\ref{table:2} (middle), we can observe an improved performance by expanding the set. It implies that the counting method based on regression labels can be incorporated into our setting and effectively boost crowd counting.

\textbf{Beyond the ``More than Twice'' Assumption.} We make a reasonable assumption that the number of people in the two images can be distinguished easily with more than twice the difference in crowd sizes between them. To explore more about the assumption, we incrementally adjust more critical times representing for difference in numbers between a pair of images by ${1N, 1.5N, 2N, 3N}$, where $2N$ represents the setting on the above assumptions in our experiments. The results are shown in Table~\ref{table:2} (right). The results indicate that the ability of annotators to recognize and compare crowd numbers at a glance can affect the performance of the proposed method with a subjective aspect.
\begin{table}[t]
	\centering
	\tabcolsep=0.10cm
	\setlength\arrayrulewidth{1.0pt}
	\caption{Impacts of the margin $\mathcal{M}$ (left), the size of counting anchor set (middle), and the crowd number difference between a ranking image pair (right). The best results are highlighted in bold.}
	\label{table:2}
	\begin{tabular}{ccc|ccc|ccc}
		\toprule
		$\mathcal{M}$& MAE & MSE & \textbf{Set size}& MAE & MSE & \textbf{Times} & MAE & MSE\\
		\midrule
		$0$ & 94.24 & 137.24 & $|B|=10$ & 115.17 & 163.75 & $1N$ & \textbf{81.25} & \textbf{126.5}\\
		$0.1$ & 95.36 & 140.95 & $|B|=30$ & 99.18 & 137.90 & $1.5N$ & 87.27 & 134.01\\
		$0.5$ & \textbf{91.66} & 138.46 & $|B|=50$ & 91.66 & 138.46 & $2N$ & 91.66 & 138.46 \\
		$1.0$ & 91.95 & \textbf{133.45} & $|B|=80$ & 87.70 & 129.04 & $3N$ & 104.06 & 144.21 \\
		$3.0$ & 95.25 & 141.27 & $|B|=150$ & \textbf{81.09} & \textbf{120.67} & & &\\
		\bottomrule
	\end{tabular}\vspace{-2mm}
\end{table}

\subsection{Challenging Experiments}
\textbf{Ranking Labels in Real-world Counting.} Unlike the crowd data in benchmarks,
the real-world crowd data is often a large quantity of unlabeled crowd images, so it is infeasible to annotate all of extracted pairs.
Especially in the mid-to-late stage of training, much abundant pairs with large numerical difference almost have no effect on loss, and just consume computing power in vain. Therefore, sparse labeling is preferable for massive, wild and unlabeled crowd data.
Denote the number of labels associated with crowd image $x_i$ as $\zeta(x_i)$, and the average number of $\zeta(x_i)$ in dataset $D$ as $\zeta(D)$. (Note that only the labeled image will be included, thus $\zeta(\cdot) \geq 1$.) In short, with sufficient training samples, $\zeta(D)$ is closer to $1$, which means the labels per image are sparse. Furthermore, online labeling strategy can work well with sparse labeling. Before the start of training, the ranking pairs are labeled under the setting, $\zeta(D) = 1$. After training for some time, $\zeta(D)$ can become larger.

Practically, to verify the proposed ranking label in real scenes, we simulate the real annotation on a fixed number of real world images. They are from JHU-CROWD++ \cite{sindagi2019pushing}, which contains 1.51 million annotated heads spanning 4,372 images, and is a challenging dataset with various scenarios.
For the sparse labeling experiment, we randomly select 2,000 images forming 1,000 ranking pairs, and the sparse labeling strategy greatly reduces the number of labels per image $\zeta(D)$, from hundreds to one. In the other two experiments, we added annotated ranking pairs besides these 2,000 images with 5,000 pairs, and 25,000 pairs to verify the impact of label intensity.

The model is trained combining ranking with free regression labels ({Ours++}) from counting anchor set which size is 50. The training is on the selected images from JHU-CROWD++, and the evaluation is on the widely-used ShanghaiTech Part A dataset. As illustrated in Table~\ref{table:4}, compared with baseline of regression on the anchor set, the model with 1,000 comparision labels achieve a satisfactory performance with a few extra annotations. Adding more ranking labels leads to a slight improvement but brings heavy annotation load.

\begin{table}[t]
	\centering
	\tabcolsep=0.10cm
	\setlength\arrayrulewidth{1.0pt}
	\caption{The effect of ranking label in the simulated real-world counting.}
	\label{table:4}
	\begin{tabular}{c||cc}
		\toprule
		\textbf{Setting} & \textbf{MAE} & \textbf{MSE} \\
		\midrule
		Baseline & 130.5 & 201.4 \\
		Ours++ (1,000 pairs) & 114.9 & 177.9 \\
		Ours++ (5,000 pairs) & 108.2 & 153.5 \\
		Ours++ (25,000 pairs) & 106.4 & 148.2 \\
		\bottomrule
	\end{tabular}\vspace{-2mm}
\end{table}

\textbf{{Cross Datasets.}}
We conduct experiments to demonstrate the generalizability of our method across different data domains. The model is trained on one dataset of a source domain and evaluated on another dataset as a target domain. The results are demonstrated in Table~\ref{table:3}. We can observe that the proposed method generalizes well to the unseen evaluation datasets. Specially, the proposed method can be comparable or even slightly better than the location-level supervised methods.
\begin{table}[t]
	\centering
	\tabcolsep=0.10cm
	\setlength\arrayrulewidth{1.0pt}
	\caption{Cross dataset experiments on the ShanghaiTech Part A, UCF-QNRF and UCF\_CC\_50 datasets for demonstrating the generalization of different methods.}
	\label{table:3}
	\begin{tabular}{c||cc|cc|cc}
		\toprule
		& \multicolumn{2}{c|}{\tiny{\textbf{ST PartA $\rightarrow$ UCF-QNRF}}} & \multicolumn{2}{c|}{\tiny{\textbf{UCF-QNRF $\rightarrow$ ST PartA}}} & \multicolumn{2}{c}{\tiny{\textbf{ST PartA $\rightarrow$ UCF\_CC\_50}}} \\
		\midrule
		\textbf{Method} & \textbf{MAE} & \textbf{MSE} & \textbf{MAE} & \textbf{MSE} & \textbf{MAE} & \textbf{MSE} \\
		\midrule
		MCNN \cite{zhang2016single} & - & - & - & - & 397.7 & 624.1  \\
		L2R \cite{liu2018leveraging} & - & - & - & - & 337.6 & 434.3  \\
		SPN \cite{xu2019learn} & 236.3 & 428.4 & 87.9 & 126.3 & 368.3 & 588.4  \\
		Ours++ & 246.1 & 485.4 & 99.2 & 141.1 & 372.6 & 593.5 \\
		Ours++(Fully) & 204.8 & 412.2 & 88.7 & 127.8 & 324.1 & 435.7 \\
		\bottomrule
	\end{tabular}\vspace{-2mm}
\end{table}

\begin{figure}[t]
	\centering
	\setlength{\tabcolsep}{1pt}
	\begin{tabular}{>{\centering\arraybackslash\hspace{0pt}}p{.49\linewidth}
			>{\centering\arraybackslash\hspace{0pt}}p{.49\linewidth}}
		\includegraphics[width=\linewidth, height = 3cm]{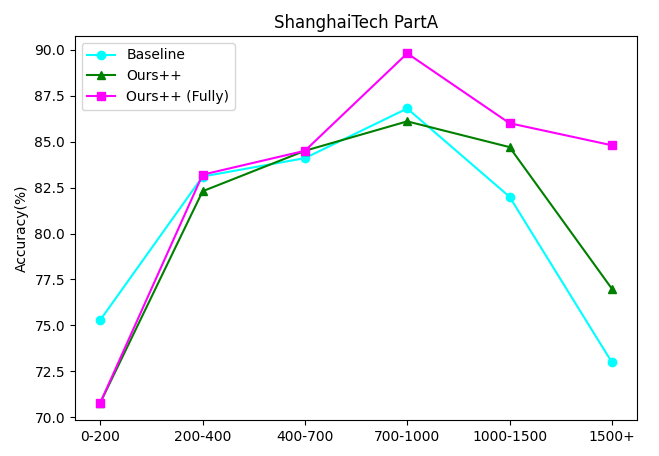} &
		\includegraphics[width=\linewidth, height = 3cm]{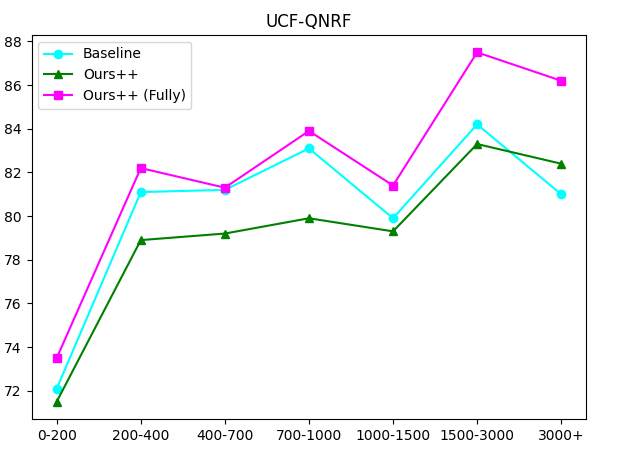}
	\end{tabular}\vspace{-1mm}
	\caption{Accuracy (\%) of each crowd number range in the proposed models and baseline.}
	\label{fig:fitting}\vspace{-2mm}
\end{figure}

\textbf{{Varied Degrees of Congestion.}}
We further evaluate the proposed methods for different degrees of crowd congestion. We observe that the comparisons of the proposed method \textit{Ours++ (Fully)} with \rll{the baseline trained on regression labels only.} It turns out that, with more people (e.g., crowd number \textgreater 1,000), \textit{Ours++} and \textit{Ours++ (Fully)} can achieve better performance. Thus, our model can well be adapted in dense scenes.
We achieve similar results on UCF-QNRF dataset as well. The proposed method \textit{Ours++ (Fully)} outperforms the baseline especially in dense scenes, as shown in Fig~\ref{fig:fitting}.
\vspace{-1mm}

\vspace{-2mm}\section{Conclusion}\label{sec:conclusion}\vspace{-1mm}
In this paper, we propose a novel weakly-supervised setting, in which we leverage the binary ranking of two images with high-contrast crowd counts as training guidance. In particular, we tailor a Siamese Ranking Network that predicts the potential scores of two images indicating the ordering of the counts.
Hence, the ultimate goal is to assign appropriate potentials for all the crowd images to ensure their orderings obey the ranking labels, and then map them to actual crowd counts.

\bibliographystyle{IEEEtran}
\bibliography{egbib}

\end{document}